# Adaptive Forecasting of Non-Stationary Nonlinear Time Series based on the Evolving Weighted Neuro-Neo-Fuzzy-ANARX-Model

**Zhengbing Hu**
School of Educational Information Technology, Central China Normal University, Wuhan, China
Email: hzb@mail.ccnu.edu.cn

**Yevgeniy V. Bodyanskiy**
Kharkiv National University of Radio Electronics, Kharkiv, Ukraine,
Email: yevgeniy.bodyanskiy@nure.ua

**Oleksii K. Tyshchenko and Olena O. Boiko**
Kharkiv National University of Radio Electronics, Kharkiv, Ukraine,
Email: lehatish@gmail.com, olena.boiko@ukr.net

*Abstract* — An evolving weighted neuro-neo-fuzzy-ANARX model and its learning procedures are introduced in the article. This system is basically used for time series forecasting. It's based on neo-fuzzy elements. This system may be considered as a pool of elements that process data in a parallel manner. The proposed evolving system may provide online processing data streams.

## I. Introduction

Mathematical forecasting of data sequences (time series) is nowadays well studied and there is a large number of publications on this topic. There are many methods for solving this task: regression, correlation, spectral analysis, exponential smoothing, etc., and more advanced intellectual systems that sometimes require rather complicated mathematical methods and high user's qualification. The problem becomes more complicated when analyzed time series are non-stationary and nonlinear and contain unknown behavior trends, quasiperiodic, stochastic and chaotic components. The best results are shown by nonlinear forecasting models based on mathematical methods of computational intelligence [1-3], and, first of all, neuro-fuzzy systems [4-5] due to their approximating and extrapolating properties, learning abilities, transparency and results' interpretability. The models to be especially noted are the so-called NARX-models [6] which have the form

$$\hat{y}(k) = f\left(y(k-1), ..., y(k-n_y), x(k-1), ..., x(k-n_x)\right) \quad (1)$$

where $\hat{y}(k)$ is an estimate of forecasted time series at discrete time $k = 1, 2, ...$; $f(\bullet)$ stands for a certain nonlinear transformation implemented by a neuro-fuzzy system, $x(k)$ is an observed exogenous factor that defines a behavior of $y(k)$. It can be noticed that popular Box–Jenkins AR-, ARX-, ARMAX-models as well as nonlinear NARMA-models can be described by the expression (1). These models have been widely studied; there are many architectures and learning algorithms that implement these models, but it is assumed that models' orders $n_y$, $n_x$ are given a priori. These orders are previously unknown in a case of structural non-stationarity for analyzed time series, and they also have to be adjusted during a learning procedure. In this case, it makes sense to use evolving connectionist systems [7-10] that adjust not only their synaptic weights and activation-membership functions, but also their architectures. There are many algorithms that implement these learning methods both in a batch mode and in a sequential mode. The problem becomes more complicated if data are fed to the system with high frequency in the form of a data stream [11]. Here, the most popular evolving systems turn out to be too cumbersome for learning and information processing in an online mode.

As an alternative, a rather simple and effective architecture can be considered. It's the so-called ANARX-model (Additive NARX) that has the form [12, 13]

$$\hat{y}(k) = f_1(y(k-1), x(k-1)) + f_2(y(k-2), x(k-2)) + \ldots$$
$$\ldots + f_n(y(k-n), x(k-n)) = \sum_{l=1}^{n} f_l(y(k-l), x(k-l)) \quad (2)$$

(here $n = \max\{n_y, n_x\}$), an original task of the forecasting system's synthesis is decomposed into many local tasks of parametric identification for node models with two input variables $y(k-l)$, $x(k-l)$, $l = 1, 2, \ldots, n, \ldots$.

Authors [12, 13] used elementary Rosenblatt perceptrons with sigmoidal activation functions as such nodes. The ANARX-model provided high forecasting quality, but generally speaking it requires a large number of nodes in its architecture [14].

Some synthesis problems of forecasting neuro-fuzzy [15] and neo-fuzzy [15-17] systems based on the ANARX-models are considered in this work. These systems avoid the above mentioned drawbacks.

Since we consider a case of stochastic nonlinear dynamic signals in this article, the basic novelty has to do with defining a model's delay order in an online mode.

## II. A NEURO-FUZZY-ANARX-MODEL

An architecture of the ANARX-model is shown in Fig.1. It is formed by two lines of time delay elements $z^{-1}$ ($z^{-1}y(k) = y(k-1)$) and $n$ nodes $N^{[l]}$ which are simultaneously learned. These nodes are tuned independently from each other. And adding new nodes or removing unnecessary ones doesn't have any influence on other neurons, i.e. the evolving process for this system is implemented by changing a number of the nodes.

It's recommended to use a neuron with two inputs (its architecture is shown in Fig.2) as a node of this system instead of the elementary Rosenblatt perceptron.

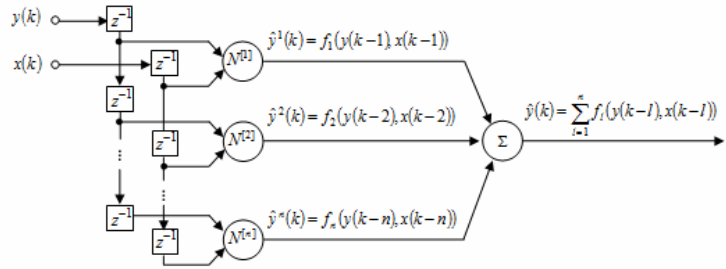

Figure 1. The ANARX-model

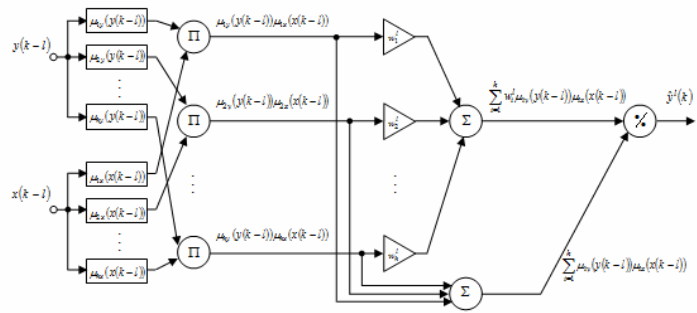

Figure 2. A neuro-fuzzy node of the ANARX-model

As one can see, this node is the Wang–Mendel neuro-fuzzy system [18] with two inputs. It possesses universal approximation capabilities and is actually the zero-order Takagi–Sugeno–Kang system [19, 20]. A two-dimensional vector of input signals $z_l(k) = (y(k-l), x(k-l))^T$ is fed to an input of the node $N^{[l]}$, $l = 1, 2, \ldots, n, \ldots$. The first layer contains $2h$ membership functions $\mu_{iy}(y(k-l))$, $\mu_{ix}(x(k-l))$ and fulfills fuzzification of the input variables by calculating membership levels $0 < \mu_{iy}(y(k-l)) \leq 1$, $0 < \mu_{ix}(x(k-l)) \leq 1$. The Gaussian functions can be used as membership functions

$$\mu_{iy}(y(k-l)) = \exp\left(-\frac{(y(k-l)-c_{iy})^2}{2\sigma_{iy}^2}\right),$$

$$\mu_{ix}(x(k-l)) = \exp\left(-\frac{(x(k-l)-c_{ix})^2}{2\sigma_{ix}^2}\right)$$

(here $c_{iy}$, $c_{ix}$ stand for parameters that define centers of these membership functions; $\sigma_{iy}$, $\sigma_{ix}$ are width parameters) or other bell-shaped functions with infinite support for avoiding "gaps" which appear in the fuzzified space.

The second layer of the node provides aggregation of the membership levels that were computed in the first layer. Outputs of the second layer form $h$ aggregated signals

$$\tilde{z}_i^l(k) = \mu_{iy}(y(k-l))\mu_{ix}(x(k-l)).$$

The third layer contains synaptic weights that are adjusted during a learning procedure. Outputs of the third layer are values

$$w_i^l \mu_{iy}(y(k-l))\mu_{ix}(x(k-l)) = w_i^l \tilde{z}_i^l(k).$$

The fourth layer is formed by two summation units and calculates sums of the output signals in the second and the third layers correspondingly. Outputs of the fourth layer are signals

$$\begin{cases} \sum_{i=1}^{h} w_i^l \mu_{iy}(y(k-l))\mu_{ix}(x(k-l)) = \sum_{i=1}^{h} w_i^l \tilde{z}_i^l(k), \\ \sum_{i=1}^{h} \mu_{iy}(y(k-l))\mu_{ix}(x(k-l)) = \sum_{i=1}^{h} \tilde{z}_i^l(k). \end{cases}$$

Defuzzification (normalization) is implemented in the fifth (output) layer. Finally, the output signal of the node $\hat{y}^l(k) = f_l(y(k-l), x(k-l))$ is computed:

$$\hat{y}^l(k) = \frac{\sum_{i=1}^{h} w_i^l \mu_{iy}(y(k-l))\mu_{ix}(x(k-l))}{\sum_{i=1}^{h} \mu_{iy}(y(k-l))\mu_{ix}(x(k-l))} =$$

$$= \sum_{i=1}^{h} w_i^l \frac{\tilde{z}_i^l(k)}{\sum_{i=1}^{h} \tilde{z}_i^l(k)} = \sum_{i=1}^{h} w_i^l \phi_i^l(k) = w^{lT}\phi^l(k)$$

where $\phi^l(k) = (\phi_1^l(k), \phi_2^l(k), ..., \phi_h^l(k))^T$, $\phi_i^l(k) = \tilde{z}_i^l(k)\left(\sum_{i=1}^{h}\tilde{z}_i^l(k)\right)^{-1}$, $w^l = (w_1^l, w_2^l, ..., w_h^l)^T$.

Considering that the output signal $\hat{y}^l(k)$ of each node depends linearly on the adjusted synaptic weights $w_i^l$, one can use conventional algorithms of adaptive linear identification [21] for their tuning which are based on the quadratic learning criterion.

If a training data set is non-stationary [22], one can use either the exponentially weighted recurrent least squares method

$$\begin{cases} w^l(k) = w^l(k-1) + \\ \dfrac{P^l(k-1)(y(k) - w^{lT}(k-1)\phi^l(k))\phi^l(k)}{\alpha + \phi^{lT}(k)P^l(k-1)\phi^l(k)}, \\ P^l(k) = P^l(k-1) - \\ \dfrac{P^l(k-1)\phi^l(k)\phi^{lT}(k)P^l(k-1)}{\alpha + \phi^{lT}(k)P^l(k-1)\phi^l(k)}, 0 < \alpha \leq 1, \end{cases} \quad (3)$$

or the Kaczmarz–Widrow–Hoff optimal gradient algorithm (in a case of the "rapid" non-stationarity)

$$w^l(k) = w^l(k-1) + \frac{y(k) - w^{lT}(k-1)\phi^l(k)}{\phi^{lT}(k)\phi^l(k)}\phi^l(k). \quad (4)$$

In fact, it is possible to tune $4h$ membership functions' parameters $c_{iy}$, $c_{ix}$, $\sigma_{iy}$, $\sigma_{ix}$ of each node, but taking into consideration the fact that the signal $\hat{y}^l(k)$ depends nonlinearly on these parameters, a learning speed can't be sufficient for non-stationary conditions in this case.

III. A NEO-FUZZY-ANARX-MODEL

If a large data set to be processed is given (within the "Big Data" conception [23] when data processing speed and computational simplicity come to the forefront, it seems reasonable to use neo-fuzzy neurons that were proposed by T. Yamakawa and his co-authors [15-17] instead of the neuro-fuzzy nodes in the ANARX-model. An architecture of the neo-fuzzy neuron as a node of the ANARX-model is shown in Fig.3. The neo-fuzzy neuron's advantages are a high learning speed, computational simplicity, good approximating properties and abilities to find a global minimum of a learning criterion in an online mode.

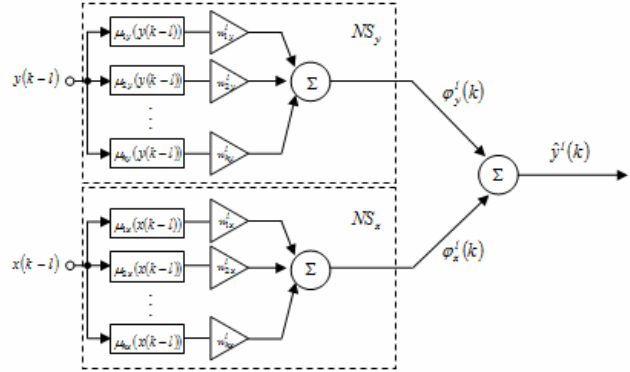

Figure 3. A neo-fuzzy node for the ANARX-model

Structural elements of the neo-fuzzy neuron are nonlinear synapses $NS_y$, $NS_x$ that implement the zero order Takagi–Sugeno fuzzy inference, but it's easy to notice that the neo-fuzzy neuron is much simpler constructively than the neuro-fuzzy node shown in Fig.2.

An output value of this node is formed with the help of input signals $y(k-l)$, $x(k-l)$

$$\hat{y}^l(k) = \phi_y^l(k) + \phi_x^l(k) = \sum_{i=1}^{h} w_{iy}^l \mu_{iy}(y(k-l)) + \sum_{i=1}^{h} w_{ix}^l \mu_{ix}(x(k-l)) \quad (5)$$

and an output signal of the ANARX-model can be written down in the form

$$\hat{y}(k) = \sum_{l=1}^{n}\left(\sum_{i=1}^{h} w_{iy}^l \mu_{iy}(y(k-l)) + \sum_{i=1}^{h} w_{ix}^l \mu_{ix}(x(k-l))\right),$$

which means that since the neo-fuzzy neuron is also an additive model [24], the ANARX-model based on the neo-fuzzy neurons is twice additive.

Triangular membership functions are usually used in the neo-fuzzy neuron. They meet the unity partitioning conditions

$$\sum_{i=1}^{h} \mu_{iy}(y(k-l)) = 1, \quad \sum_{i=1}^{h} \mu_{ix}(x(k-l)) = 1,$$

which make it possible to simplify the node's architecture by excluding the normalization layer from it.

It was proposed to use B-splines as membership functions for the neo-fuzzy neuron [25]. They provide a higher approximation quality and also meet the unity partitioning conditions. A B-spline of the $q$-th order can be written down:

$$\mu_{iy}^{q}\left(y(k-l)\right) = \begin{cases} \begin{rcases} 1, \text{ if } c_{iy} \leq y(k-l) < c_{i+1,y}, \\ 0 \text{ otherwise} \end{rcases} \text{ for } q = 1 \\ \\ \dfrac{y(k-l) - c_{iy}}{c_{i+q-1,y} - c_{iy}} \mu_{iy}^{q-1}\left(y(k-l)\right) + \\ \dfrac{c_{i+q,y} - y(k-l)}{c_{i+q,y} - c_{i+1,y}} \mu_{i+1,y}^{q-1}\left(y(k-l)\right), \text{ for } q > 1 \\ \\ i = 1, \ldots, h-q, \end{cases}$$

$$\mu_{ix}^{q}\left(x(k-l)\right) = \begin{cases} \begin{rcases} 1, \text{ if } c_{ix} \leq x(k-l) < c_{i+1,x}, \\ 0 \text{ otherwise} \end{rcases} \text{ for } q = 1 \\ \\ \dfrac{x(k-l) - c_{ix}}{c_{i+q-1,x} - c_{ix}} \mu_{ix}^{q-1}\left(x(k-l)\right) + \\ \dfrac{c_{i+q,x} - x(k-l)}{c_{i+q,x} - c_{i+1,x}} \mu_{i+1,x}^{q-1}\left(x(k-l)\right), \text{ for } q > 1 \\ \\ i = 1, \ldots, h-q. \end{cases}$$

It should be mentioned that B-splines are a sort of generalized membership functions: when $q = 2$ one gets traditional triangular membership functions; when $q = 4$ one gets cubic splines, etc.

Introducing a vector of variables $\phi^{l}(k) = \left(\mu_{1y}\left(y(k-l)\right), \ldots, \mu_{hy}\left(y(k-l)\right), \mu_{1x}\left(x(k-l)\right), \ldots, \mu_{hx}\left(x(k-l)\right)\right)^{T}$, $w^{l} = \left(w_{1y}^{l}, \ldots, w_{hy}^{l}, w_{1x}^{l}, \ldots, w_{hx}^{l}\right)^{T}$, the expression (5) can be rewritten in the form

$$\hat{y}^{l}(k) = w^{lT}\phi^{l}(k).$$

Either the algorithms (3), (4) or the procedure [26]

$$\begin{cases} w^{l}(k) = w^{l}(k-1) + r_{l}^{-1}(k)\left(y(k) - w^{lT}(k-1)\phi^{l}(k)\right)\phi^{l}(k) \\ r_{l}(k) = \alpha r_{l}(k-1) + \phi^{lT}(k)\phi^{l}(k), 0 \leq \alpha \leq 1, \end{cases} \quad (6)$$

can be used for the neo-fuzzy neuron's learning. This procedure possesses both filtering and tracking properties. It should also be noticed that when $\alpha = 1$ the equation (6) coincides completely with the Kaczmarz–Widrow–Hoff optimal algorithm (4).

Evolving systems based on the neo-fuzzy neurons demonstrated its effectiveness for solving different tasks (especially forecasting tasks [25-30]). The twice additive system is the most appropriate choice for data processing in Data Stream Mining tasks [11] from a point of view of implementation simplicity and a processing speed.

## IV. A WEIGHTED NEURO-NEO-FUZZY-ANARX-MODEL

Considering that every node $N^{[l]}$ of the ANARX-model is tuned independently from each other and represents an individual neuro(neo)-fuzzy system, it is possible to use a combination of neural networks' ensembles [31] in order to improve a forecasting quality. This approach results in an architecture of a weighted neuro-neo-fuzzy-ANARX-model (Fig.4).

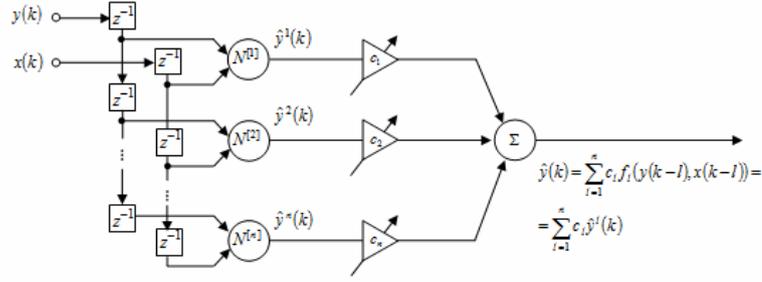

Figure 4. The weighted ANARX-model

An output signal of the system can be written in the form

$$\hat{y}(k) = \sum_{l=1}^{n} c_l \hat{y}^l(k) = c^T \hat{\bar{y}}(k)$$

where $\hat{\bar{y}}(k) = \left(\hat{y}^1(k), \hat{y}^2(k), ..., \hat{y}^n(k)\right)^T$, $c = (c_1, c_2, ..., c_n)^T$ is a vector of adjusted weight coefficients that define proximity of the signals $\hat{y}^l(k)$ to the forecasted process $y(k)$ and meet the unbiasedness condition

$$\sum_{l=1}^{n} c_l = c^T I_n = 1$$

where $I_n$ is a $(n \times 1)$-vector of unities.

The method of undetermined Lagrange multipliers can be used for finding the vector $c$ in a batch mode. So, a sequence of forecasting errors

$$v(k) = y(k) - \hat{y}(k) = y(k) - c^T \hat{\bar{y}}(k) = c^T I_n y(k) - c^T \hat{\bar{y}}(k) = $$
$$= c^T \left(I_n y(k) - \hat{\bar{y}}(k)\right) = c^T V(k),$$

the Lagrange function

$$L(c, \lambda) = \sum_k c^T V(k) V^T(k) c + \lambda \left(c^T I_n - 1\right) = $$
$$= c^T R c + \lambda \left(c^T I_n - 1\right) \quad (7)$$

(here $\lambda$ is an undetermined Lagrange multiplier, $R = \sum_k V(k) V^T(k)$ stands for an error correlation matrix) and the Karush–Kuhn–Tucker system of equations

$$\begin{cases} \nabla_c L(c, \lambda) = 2Rc + \lambda I_n = \vec{0}, \\ \dfrac{\partial L}{\partial \lambda} = c^T I_n - 1 = 0 \end{cases} \quad (8)$$

are introduced. A solution of the system (8) can be written down in the form

$$\begin{cases} c = R^{-1} I_n \left(I_n^T R^{-1} I_n\right)^{-1}, \\ \lambda = -2 I_n^T R^{-1} I_n \end{cases} \quad (9)$$

where a value of the Lagrange function (7) at the saddle point is

$$L^*(c, \lambda) = \left(I_n^T R^{-1} I_n\right)^{-1}.$$

An implementation procedure of the algorithm (9) can meet some problems in a case of data processing in an online mode and high level of correlation between the signals $\hat{y}^l(k)$ that leads to the ill-conditioned matrix $R$ which has to be inverted at every time step $k$.

The Lagrange function (7) can be written down in the form

$$L(c, \lambda) = \sum_k \left(y(k) - c^T \hat{\bar{y}}(k)\right)^2 + \lambda \left(c^T I_n - 1\right)$$

and a gradient algorithm for finding its saddle point based on the Arrow–Hurwicz procedure [27, 32] can be written in the form

$$\begin{cases} c(k) = c(k-1) - \eta_c(k) \nabla_c L(c, \lambda), \\ \lambda(k) = \lambda(k-1) + \eta_\lambda(k) \dfrac{\partial L(c, \lambda)}{\partial \lambda} \end{cases}$$

or

$$\begin{cases} c(k) = c(k-1) + \eta_c(k) \times \\ \times \left(2\left(y(k) - c^T(k-1)\hat{\bar{y}}(k)\right)\hat{\bar{y}}(k) - \lambda(k-1)I_n\right) = \\ = c(k-1) + \eta_c(k)\left(2v(k)\hat{\bar{y}}(k) - \lambda(k-1)I_n\right), \\ \\ \lambda(k) = \lambda(k-1) + \eta_\lambda(k)\left(c^T(k)I_n - 1\right) \end{cases} \quad (10)$$

where $\eta_c(k)$, $\eta_\lambda(k)$ are learning rate parameters.

The Arrow–Hurwicz procedure converges to the saddle point with rather general assumptions on the values $\eta_c(k)$, $\eta_\lambda(k)$, but these parameters can be optimized to speed up the learning process as that is particularly important in Data Stream Mining tasks.

That's why the first ratio in the equation (10) should be multiplied by $\hat{\bar{y}}^T(k)$

$$\hat{\bar{y}}^T(k)c(k) = \hat{\bar{y}}^T c(k-1) + \\ + \eta_c(k)\left(2v(k)\left\|\hat{\bar{y}}(k)\right\|^2 - \lambda(k-1)\hat{\bar{y}}^T I_n\right)$$

and an additional function that characterizes criterial convergence is introduced

$$\left(y(k) - \hat{\bar{y}}^T(k)c(k)\right)^2 = v^2(k) - \\ -2\eta_c(k)v(k)\left(2v(k)\left\|\hat{\bar{y}}(k)\right\|^2 - \lambda(k-1)\hat{\bar{y}}^T I_n\right) + \\ +\eta_c^2(k)\left(2v(k)\left\|\hat{\bar{y}}(k)\right\|^2 - \lambda(k-1)\hat{\bar{y}}^T I_n\right)^2$$

A solution for the differential equation

$$\frac{\partial\left(y(k) - \hat{\bar{y}}^T(k)c(k)\right)^2}{\partial \eta_c(k)} = \\ -2v(k)\left(2v(k)\left\|\hat{\bar{y}}(k)\right\|^2 - \lambda(k-1)\hat{\bar{y}}^T I_n\right) + \\ +2\eta_c(k)\left(2v(k)\left\|\hat{\bar{y}}(k)\right\|^2 - \lambda(k-1)\hat{\bar{y}}^T I_n\right)^2 = 0$$

gives the optimal learning rate values $\eta_c(k)$ in the form

$$\eta_c(k) = \frac{v(k)}{2v(k)\left\|\hat{\bar{y}}(k)\right\|^2 - \lambda(k-1)\hat{\bar{y}}^T I_n}.$$

Applying this value to the expression (10), final equations can be written down

$$\begin{cases} c(k) = c(k-1) + \dfrac{v(k)\left(2v(k)\hat{\bar{y}}(k) - \lambda(k-1)I_n\right)}{2v(k)\left\|\hat{\bar{y}}(k)\right\|^2 - \lambda(k-1)\hat{\bar{y}}^T I_n}, \\ \lambda(k) = \lambda(k-1) + \eta_\lambda(k)\left(c^T(k)I_n - 1\right). \end{cases} \quad (11)$$

It's easy to notice that when $\lambda(k-1) = 0$ the procedure (11) coincides with the Kaczmarz–Widrow–Hoff algorithm (4).

V. Experiments

In order to prove the effectiveness of the proposed system, a number of experiments should be carried out.

*A. Electricity Demand*

This data set describes 15 minutes averaged values of power demand in the full year 1997. Generally speaking, this data set contains 15000 points but we took only 5000 points for the experiment. 3000 points were selected for a training stage and 2000 points were used for testing. Prediction results of the ANARX and weighted ANARX systems are in Fig.5 and Fig.6 correspondingly (signal values are marked with a blue color; prediction values are marked with a magenta color; and prediction errors are marked with a grey line).

We used for comparison multilayer perceptrons (MLPs), radial-basis function neural networks (RBFNs), ANFIS and two proposed systems ANARX and weighted ANARX (both based on neo-fuzzy nodes). Since MLP can't work in an online mode, it processed data in two different modes. It had just one epoch in the first case (something similar to an online case), and it had 5 epochs in the second case (the MLP architecture had almost the same number of adjustable parameters when compared to the proposed systems). A number of MLP inputs was equal to 4 and a number of hidden nodes was equal to 7 in both cases. A total number of parameters to be tuned was 43 in both modes. It took about two times more time to compute the result in the second case but prediction quality was like almost two times higher. MLP (case 2) demonstrated the best result in this experiment.

Speaking of RBFNs, we also had two cases. The first-case RBFN was taken really close in the sense of parameters' number to our systems and the second-case RBFN's architecture was chosen to show the best performance. In the first case, RBFN had 3 inputs and 7 kernel functions. In the second case, it had 3 inputs as well but 12 kernel functions which generally led to higher prediction quality (+30% precision compared to RBFN in the first case) but took longer to compute the result. A number of parameters to be tuned was 36 in the first case and 61 in the second case.

ANFIS showed one of the best prediction qualities in this experiment. It had 4 inputs, 55 nodes and it was processing data during 5 epochs. It contained 80 parameters to be tuned.

The proposed ANARX system based on neo-fuzzy nodes had 2 inputs, 2 nodes, 9 membership functions, and its $\alpha$ parameter was equal to 0.62. This system had 37 parameters. Its prediction quality was rather high and it was definitely one of the fastest system on this data set. We should also notice that we used B-splines (q=2, which means that we used triangular membership functions) as membership functions for the proposed systems.

The proposed weighted ANARX system based on neo-fuzzy nodes had 2 inputs, 2 nodes, 8 membership functions; its $\alpha$ parameter was equal to 0.9. It had 37 adjustable parameters. This system demonstrated better performance when compared to ANARX and the fastest results.

TABLE I. COMPARISON OF THE SYSTEMS' RESULTS

| Systems | Parameters to be tuned | RMSE (training) | RMSE (test) | Time, s |
|---|---|---|---|---|
| MLP (case 1) | 43 | 0.0600 | 0.0700 | 0.4063 |
| MLP (case 2) | 43 | 0.0245 | 0.0381 | 0.9219 |
| RBFN (case 1) | 36 | 0.0681 | 0.0832 | 0.6562 |
| RBFN (case 2) | 61 | 0.0463 | 0.0604 | 1.0250 |
| ANFIS | 80 | 0.0237 | 0.0396 | 0.7031 |
| ANARX (neo-fuzzy nodes) | 37 | 0.0903 | 0.0922 | 0.4300 |
| Weighted ANARX (neo-fuzzy nodes) | 36 | 0.0427 | 0.0573 | 0.3750 |

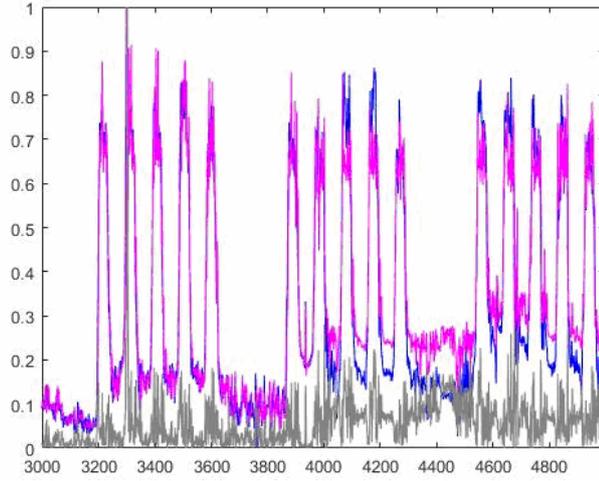
Figure 5. Identification of a nonlinear system. Prediction results of the ANARX system

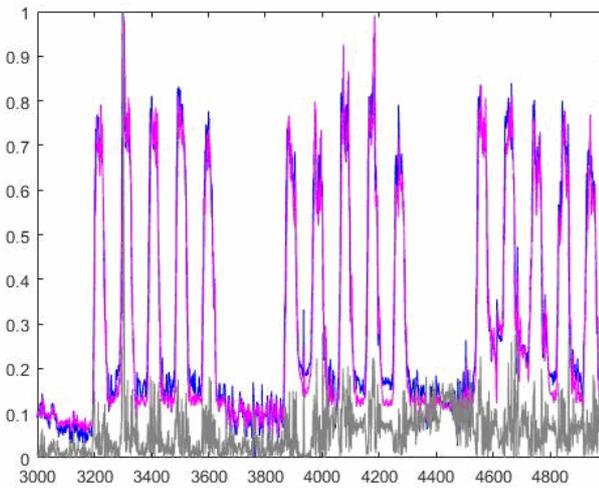
Figure 6. Identification of a nonlinear system. Prediction results of the weighted ANARX system

*B. Monthly sunspot number*

This data set was taken from datamarket.com. The data set describes monthly sunspot number in Zurich. It was collected between 1749 and 1983. It contains 2820 points. 2256 points were selected for a training stage and 564 points were used for testing. Prediction results of the ANARX and weighted ANARX systems are in Fig.7 and Fig.8 correspondingly (signal values are marked with a blue color; prediction values are marked with a magenta color; and prediction errors are marked with a grey line).

We used a set of systems which is similar to the previous experiment to compare results. MLP_1 (MLP for the first case) had 3 inputs, 6 hidden nodes and it was processing data during just one epoch. MLP_2 basically had the same parameter settings except the fact that it was processing data during 5 epochs. Both MLP systems had 31 parameters to be tuned. MLP_1 was 2 times faster than MLP_2 but its prediction quality was also almost 2 times worse.

Let's denote our RBFN architectures as RBFN_1 (RBFN for the first case) and RBFN_2 (RBFN for the second case). Both of them had 3 inputs. RBFN_1 had 4 kernel functions unlike RBFN_2 which had 19 hidden nodes. A number of parameters to be tuned was 21 in the first case and 96 in the second case. RBFN_2 showed a better prediction quality but it was much slower than RBFN_1.

ANFIS had 4 inputs and 55 nodes. It was processing data during 3 epochs. It had 80 adjustable parameters.

The proposed ANARX system based on neo-fuzzy nodes had 2 inputs, 2 nodes, 4 membership functions, and its $\alpha$ parameter was equal to 0.9. This system contained 17 parameters. Its prediction quality was rather high and it was definitely one of the fastest systems on this data set.

The proposed weighted ANARX system based on neo-fuzzy nodes had 2 inputs, 2 nodes, 4 membership functions; its $\alpha$ parameter was equal to 0.9. This system contained 20 parameters to be tuned. This system demonstrated better performance when compared to ANARX and one of the fastest results.

TABLE II. COMPARISON OF THE SYSTEMS' RESULTS

| Systems | Parameters | RMSE | RMSE | Time, s |
|---------|------------|------|------|---------|

|  | to be tuned | (training) | (test) |  |
|---|---|---|---|---|
| MLP (case 1) | 31 | 0.1058 | 0.1407 | 0.2813 |
| MLP (case 2) | 31 | 0.0600 | 0.0808 | 0.5938 |
| RBFN (case 1) | 21 | 0.1066 | 0.2155 | 0.2219 |
| RBFN (case 2) | 96 | 0.0702 | 0.1638 | 0.9156 |
| ANFIS | 80 | 0.0559 | 0.1965 | 0.8906 |
| ANARX (neo-fuzzy nodes) | 17 | 0.1297 | 0.1350 | 0.2252 |
| Weighted ANARX (neo-fuzzy nodes) | 20 | 0.0784 | 0.1081 | 0.2981 |

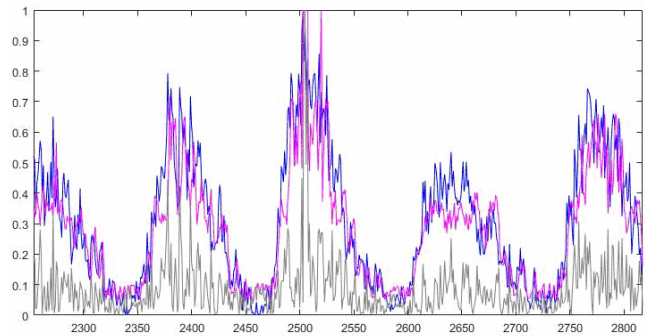

Figure 7. Identification of a nonlinear system. Prediction results of the ANARX system

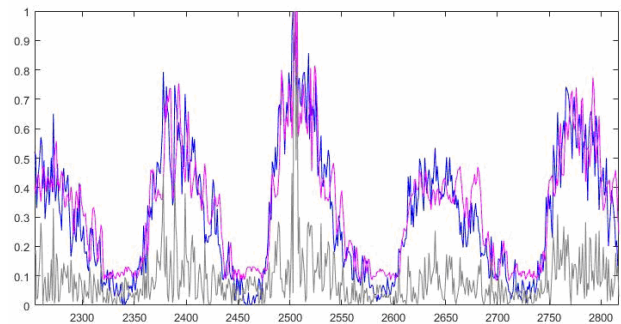

Figure 8. Identification of a nonlinear system. Prediction results of the weighted ANARX system

## VI. CONCLUSION

The evolving forecasting weighted neuro-neo-fuzzy-twice additive model and its learning procedures are proposed in the paper. This system can be used for non-stationary nonlinear stochastic and chaotic time series' forecasting where time series are processed in an online mode under the parametric and structural uncertainty. The proposed weighted ANARX-model is rather simple from a computational point of view and provides fast data stream processing in an online mode.

So, the proposed evolving forecasting model has demonstrated its efficiency for solving real-world tasks. A number of experiments has been performed to show high efficiency of the proposed neuro-neo-fuzzy system.

## ACKNOWLEDGMENT

The authors would like to thank anonymous reviewers for their careful reading of this paper and for their helpful comments.

This scientific work was supported by RAMECS and CCNU16A02015.